\documentclass[graybox]{svmult}

\usepackage{mathptmx}       
\usepackage{helvet}         
\usepackage{courier}        
\usepackage{type1cm}        
\usepackage{makeidx}         
\usepackage{graphicx}        
\usepackage{multicol}        
\usepackage[bottom]{footmisc}

\usepackage{biblatex}
\usepackage{hyperref}
\usepackage{comment}
\usepackage{adjustbox}
\usepackage{booktabs}
\usepackage{geometry}
\usepackage{array}
\usepackage{caption}
\usepackage{float}
\usepackage{amsmath}
\usepackage{amsfonts}
\usepackage{xcolor}

\makeindex             

\begin{document}

\title*{A Unified Approach to Emotion Detection and Task-Oriented Dialogue Modeling}

\titlerunning{A Unified Approach to Emotion Detection and Task-Oriented Dialogue Modeling}
\author{Armand Stricker, Patrick Paroubek}
\institute{Armand Stricker, Patrick Paroubek \at Université Paris-Saclay, CNRS, Laboratoire Interdisciplinaire des Sciences du Numérique, 91400, Orsay, France \email{firstname.lastname@lisn.upsaclay.fr}}
%
%

\maketitle

\abstract*{}


\abstract{In current task-oriented dialogue (TOD) systems, text-based user emotion detection (ED) is often overlooked or is typically treated as a separate and independent task, requiring additional training. In contrast, our work demonstrates that seamlessly unifying ED and TOD modeling brings about  mutual benefits, and is therefore an alternative to be considered. Our method consists in augmenting SimpleToD, an end-to-end TOD system, by extending belief state tracking to include ED, relying on a single language model. We evaluate our approach using GPT-2 and Llama-2 on the EmoWOZ benchmark, a version of MultiWOZ annotated with emotions. Our results reveal a general increase in performance for ED and task results. Our findings also indicate that user emotions provide useful contextual conditioning for system responses, and can be leveraged to further refine responses in terms of empathy.}


\section{Introduction}
\label{sec:1}

Emotional user utterances frequently occur during interactions with dialogue systems \cite{reeves1996media}. In open-domain dialogues, users directly share personal and emotional experiences \cite{li-etal-2017-dailydialog}. In task-oriented dialogues (TODs), emotions are closely related to task progression. They usually become apparent as the task unfolds, and are contingent on the user's expectations being met \cite{feng-etal-2022-emowoz}. Detecting these emotions explicitly in either scenario is beneficial for several reasons: it can help with reviewing chat logs after the exchange concludes or with adjusting responses during the exchange to better align with the user's emotional state \cite{rashkin-etal-2019-towards}. In TODs, these more empathetic responses have demonstrated their effectiveness in compensating for system errors \cite{i-feel-you-2013}, creating the impression of a more capable system.


Existing text-based approaches to ED in TODs either require a dedicated, specifically trained component \cite{feng-etal-2022-emowoz,feng-etal-2023-chatter} or assume \textit{implicit} ED. This assumption is due to systems being commonly trained to replicate human expert responses \cite{xu2017new, ham-etal-2020-end} which inherently convey empathy, as needed, and therefore implicit ED. In contrast, we propose an approach that eliminates the need for training an additional component while \textit{explicitly} modeling user emotions, applicable in settings where emotion annotations are available. Our method involves predicting user emotions within a sequence of task-related components, treating ED as an extension of belief state tracking. Our approach is end-to-end and relies on a single language model, optimized in a unified fashion. Specifically, we build upon SimpleTOD \cite{hosseini-asl_simple_2020}, a popular end-to-end TOD approach trained on GPT-2 \cite{radford2019gpt2} (Sec. \ref{para: method}). Recognizing the success of large language models (LLMs) \cite{brown2020language}, we also fine-tune a state-of-the-art  LLM, Llama-2 \cite{touvron2023llama2}, using LoRA \cite{hu2021lora}, a parameter-efficient fine-tuning technique (Sec. \ref{para: method}).  

Moreover, we show that when considering Llama-2, ED predictions can be readily used to more explicitly condition and refine system responses. Capitalizing on the fact that LoRA weights are optimized \textit{on top} of Llama-2's fixed parameters, we dynamically disable the LoRA weights after generating TOD predictions to harness the LLM's in-context learning ability. We refine responses without further training, using a chain-of-thought prompting approach \cite{wei2022chain} (Sec. \ref{para: method}). 

We evaluate our method on the text-based EmoWOZ benchmark \cite{feng-etal-2022-emowoz}, the only benchmark to provide emotion annotations on MultiWOZ \cite{budzianowski-etal-2018-multiwoz}. The authors propose several competitive ED-only baselines, with ContextBERT \cite{feng-etal-2022-emowoz} being the most successful and taking the entire dialogue history as input.

Results demonstrate that our unified approach generally leads to improved ED outcomes compared to ContextBERT, while also yielding comparable to superior task-oriented results compared with SimpleToD.  Human evaluation of outputs produced by Llama-2 indicates that both refined responses and those generated by our modified pipeline are preferred by annotators when compared with SimpleToD. More specifically, refined responses display more empathy and engagingness. Our findings encourage a more direct integration of ED into the task-oriented framework, moving beyond its treatment as an independent task. Our experimental code can be found on GitHub\footnote{\url{https://github.com/armandstrickernlp/Emo-TOD}}.

\section{Method and Experimental Setup} \label{sec: method}

\paragraph{\textbf{The EmoWOZ Dataset}}\label{par: Dataset} 
This is the only text-based dataset offering emotion annotations for a large and popular corpus of 10,000 TODs. It focuses on 7 user emotions: \textbf{neutral, excited, fearful, satisfied, dissatisfied, apologetic}, and \textbf{abusive}. These follow the OCC model (Ortony, Clore and Collins) \cite{ortony_clore_collins_1988} and are based on 3 possible emotion elicitors which include the \textbf{operator} (or system), the \textbf{user}, and \textbf{events}. This emotion schema is highly specific to TODs and makes it possible to distinguish between emotions caused by the system and those provoked by external factors, independent of the system’s performance.

\begin{itemize}
    \item Positive emotions caused by the \textbf{operator} relate to the system successfully completing the task and are classified as \textit{satisfied, liking, appreciative}. Negative emotions caused by the operator (e.g. proposing the wrong type of restaurant) are labeled as \textit{dissatisfied, disliking} or as \textit{abusive} in some cases.

    \item A single emotion is attributed to the \textbf{user}, and is labeled \textit{apologetic}. It is employed to denote negative emotions that arise when the user's own actions or queries lead to confusion for the operator.

    \item Emotions elicited by \textbf{events} are tagged with the label \textit{fearful, sad, disappointed} when negative (e.g. the user's favorite restaurant is unavailable), and with the label \textit{excited, happy, anticipating} otherwise (e.g. the user eagerly anticipates an upcoming trip).

    \item The default label when no emotion is expressed is \textit{neutral}.   
\end{itemize}

Examples of user utterances and their associated emotion label can be found in Appendix 4. Dataset statistics are shown in Table \ref{table:dataset}. Certain emotions are notably rare and make this a challenging benchmark for ED in TODs.  

\begin{table}[ht]
\centering
\scalebox{1.1}{
\begin{tabular}{c|c|c|c|c|c|c|c}
\toprule
 & Neut. & Fear. & Diss. & Apol. & Abus. & Exci. & Sati. \\
\midrule
proportion & 71.9\% & 0.5\% & 1.3\% & 1.2\% & 0.1\% & 1.2\% & 23.8\% \\
count & 51,426 & 381 & 914 & 838 & 44 & 860 & 17,061 \\
\bottomrule
\end{tabular}
}
\caption{\small(EmoWOZ label statistics.  Emotions are \textbf{Neut}ral; \textbf{Fear}ful, sad, disappointed; \textbf{Diss}atisfied, disliking; \textbf{Apol}ogetic; \textbf{Abus}ive; \textbf{Exci}ted, happy, anticipating; \textbf{Sati}sfied, liking, appreciative.}
\label{table:dataset}
\end{table}


\paragraph{\textbf{Method}} \label{para: method}
Our approach builds upon SimpleToD \cite{hosseini-asl_simple_2020}, a popular end-to-end strategy that employs a single language model, similar to other recent end-to-end TOD systems \cite{ham-etal-2020-end, yang2021ubar}. In training, the model is exposed to instances where Context (all preceding turns $C_{t} = [ U_{0}, S_{0}, ... ,U_{t} ]$), Belief State ($B_{t}$), Dialogue Acts ($A_{t}$) and System Response ($S_{t}$) are concatenated into a single text sequence.  To ensure database adaptability, responses are additionally delexicalized. During inference, only the Context $C_{t}$ is passed as input, and the model is expected to generate the rest of the sequence. Moreover, the predicted belief state is used to query the database to retrieve values with which to replace delexicalized placeholders. This pipeline can be readily augmented, and previous studies have in fact done so with snippets of knowledge \cite{chen-etal-2022-ketod} and chitchat \cite{sun-etal-2021-adding}.

In our work, we enhance SimpleToD by \textbf{extending the belief state}, introducing an additional ED component into the pipeline. We cast both ED and TOD as tasks to be learned in a unified way with a single language modeling objective, maximizing next token probability given a sequence of input tokens: $L = \sum_i log P (t_i|t_{<i})$. 


We propose two variants of our approach. The \textbf{EMO} variant is trained on sequences which take the form $\color{black}[C_{t}, B_{t}, \color{blue}{E}_{t}\color{black}, A_{t}, S_{t}]$ , where $\color{blue}{E}_{t}\color{black}$ is the emotion label associated with the user utterance $U_{t}$.  The \textbf{PREV} variant is trained on sequences made up of  $[\color{blue}C_{t}^{+}\color{black}, B_{t}, \color{blue}{E}_{t}\color{black}, A_{t}, S_{t}]$ with $\color{blue}C_{t}^{+}\color{black}=[(U_{0}, E_{0}), S_{0}, ... , U_{t}]$. The notable difference between both variants is that $\color{blue}C_{t}^{+}\color{black}$ concatenates previous user utterances (excluding $U_{t}$) with their corresponding emotion labels. This is meant to help the model pick up on potential emotional progression patterns in the dialogues. 

\begin{figure}[htb]
\centering
\includegraphics[width=15cm]{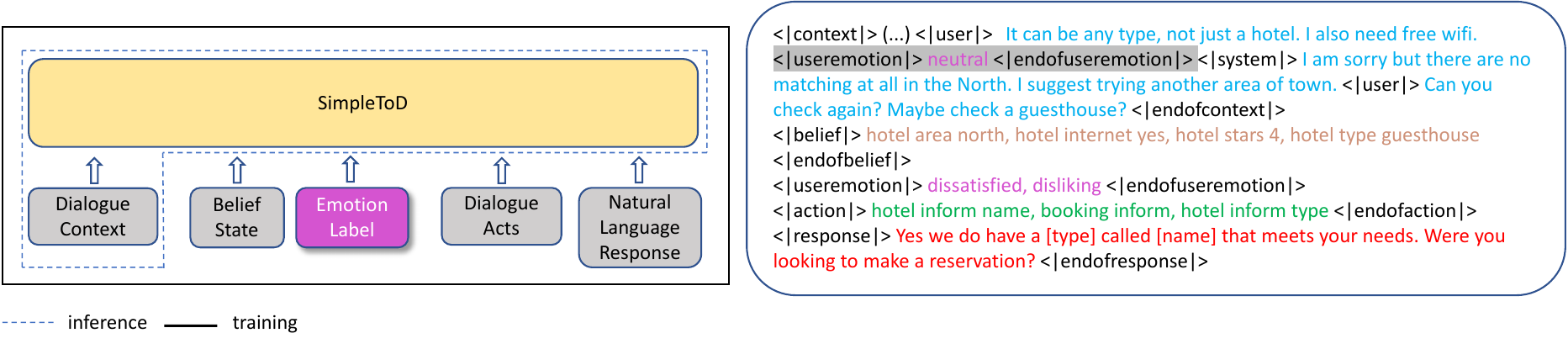}
\caption{\small (left) We expand the belief state to add emotion detection to the standard pipeline. During inference, only the dialogue Context is passed to SimpleToD and each component is generated in an autoregressive manner. (right) A turn from dialogue pmul0450, formatted for our method. Previous user emotions, highlighted in gray, are added to the Context for the PREV variant. Each component is delineated by special tokens, which helps the model in its predictions, and delexicalized placeholders in the response are in brackets. }
\label{fig: method}
\end{figure}

During inference, both variants are expected to generate each component given only an input Context. In the case of PREV, the generated emotion label is inserted back into the Context to condition the next turn's predictions. We show an overview of the approach as well as an example of the input format in Figure \ref{fig: method}.

The models used for our experiments include GPT-2\footnote{\url{https://huggingface.co/gpt2}} \cite{radford2019gpt2}, typically used with SimpleToD, and the more recent, state-of-the-art Llama-2-chat\footnote{\url{https://huggingface.co/meta-llama/Llama-2-7b-chat-hf}}\cite{touvron2023llama2}.  This is an open-source LLM trained on data from publicly accessible sources which can undergo further fine-tuning or be prompted directly with instructions and/or few-shot examples. To adapt models to our task, we fully fine-tune GPT-2 and employ LoRA fine-tuning for Llama-2, preserving a significant portion of the pre-trained weights. This approach enhances regularization, generally leading to improved generalization compared to a fully fine-tuned alternative. Training details can be found in Appendix 2.

When \textbf{refining responses}, we focus on the PREV model's predictions. Once these have been generated, we can disable the LoRA weights and, with no additional training, prompt the model using a few-shot, chain-of-thought \cite{wei2022chain} approach (\textbf{REFINE}).  This consists in creating demonstrations which include a reasoning step to prime the LLM's output.  Each few-shot demonstration consists in \textbf{a)} the dialogue Context, \textbf{b)} the emotion to be predicted, \textbf{c)} the response to be generated, \textbf{d)} a chain-of-thought reasoning step suggesting which behavior to display. Finally, it also contains \textbf{e)} an emotion-aware snippet that can be \textit{prepended} to the previously generated response. Compared with refining the entire response, prompting the model to produce a prependable snippet mitigates task-related errors due to substantial modifications made by the LLM. We incorporate specific instructions into the prompt and construct one exemplar per emotion, picking examples from the training set. The demonstration snippets we design are meant to convey remorse, reassurance, appeasement, or enthusiasm, depending on the user's emotion. Appendix 3 shows an illustrative example. When generating snippets, we filter those which are overly similar to the generated response to avoid redundancy, based on Levenshtein distance. Empirically, we find that filtering out snippets with a similarity ratio $\geq$ 50\% is effective.

\paragraph{\textbf{Automatic Metrics}}
For ED, we adhere to the EmoWOZ metrics, presenting \textbf{F1 scores} for each emotion alongside the \textbf{macro} and \textbf{weighted} averages excluding the \textit{neutral} class.  For end-to-end TOD, we employ the standard MultiWOZ metrics. The \textbf{joint goal accuracy} (JGA) reflects the proportion of turns where the predicted user constraints exactly match the reference ones. The \textbf{inform rate} evaluates the system’s capacity to provide the right type of entities from the database, given the user's constraints, and the \textbf{success rate} assesses how effectively the system delivers requested attributes like phone numbers or booking references. For a more comprehensive understanding of these metrics, we direct readers to the MultiWOZ paper \cite{budzianowski-etal-2018-multiwoz}.  For response quality, we use the \textbf{BLEU} \cite{papineni-etal-2002-bleu} score. For response diversity, we analyze the \textbf{Conditional Bigram Entropy} (CBE) and the count of \textbf{unique trigrams} to gauge the richness of vocabulary and phrasing \cite{nekvinda-dusek-2021-shades}.

\section{Results and Discussion} \label{sec: results}

\begin{table*}[ht]
\centering
\resizebox{11cm}{!}{%
\begin{tabular}{l c c c c c c c c c}
\toprule
Model & Neut. & Fear. & Diss. & Apol. & Abus. & Exci. & Sat. & Macro & Weigh. \\
\midrule
ContextBERT & 95.1 & 35.7 & 36.4 & 70.3 & 19.4 & 34.1 & 90.0 & 47.7 & 83.8 \\
\midrule
EMO-gpt& 95.5 & 38.7 & 37.3  & 71.1 & 27.4 & 40.8  & 90.7 & 51.0 & 84.8 \\
PREV-gpt& \textbf{95.6}* & 21.5 & \textbf{40.5}* & 73.4* & 27.9 & 41.9 & \textbf{91.1}* & 49.4 & \textbf{85.3} \\
\midrule
EMO-llama& 95.4 & 51.7* & 34.5 & 71.0 & 21.3 & 39.8 & 90.5 & 51.5 & 84.6 \\
PREV-llama& \textbf{95.6}* & \textbf{55.2} & 37.9 & \textbf{74.2} & \textbf{36.7}* & \textbf{44.0}* & \textbf{91.1}* & \textbf{55.4}* & 85.2* \\
\bottomrule
\end{tabular}
}
\caption{\small Mean F1-scores (5 seeds) for individual emotion labels, as well as macro and weighted averages (excluding \textit{neutral}). In each column, best values are in bold and * indicates statistical significance ($p<$0.05, paired t-test) between best and second best values. If no statistical significance is observed, we compare second and third best values.}
\label{tab:emo_res}
\end{table*}

With regards to \textbf{Emotion Detection} (Table \ref{tab:emo_res}), jointly learning ED and TOD modeling yields improvements over ContextBert, EmoWOZ's most effective ED-only model, with the exception of one result (\textit{fearful}, PREV-gpt).  

For EMO-gpt, a model comparable in size to the baseline, improvements can be seen in macro and weighted averages. This indicates that learning to predict task-related components, such as the belief state and dialogue acts, provides useful information for ED. This observation aligns with the idea that emotions in TODs are closely linked to task progression, making task-related information beneficial. As for the PREV-gpt variant, the addition of predicted emotions to the Context introduces noise for \textit{fearful} but proves beneficial for \textit{dissatisfied}. In cases where the weighted average matters most, this PREV variant is preferable. 

For Llama-2, the PREV variant shows a clearer benefit. This model performs best overall, except for the \textit{dissatisfied} emotion.  By examining user utterances with this particular emotion, we find that they appear highly contextual and generally lack explicit signs of dissatisfaction, a common trait when identifying emotions from text only \cite{multimodal_emo_rec}. The regularization from LoRA in this case is a drawback  compared to full fine-tuning, as annotations for this emotion are dataset-specific.  However, for less common emotions characterized by more explicit expressions, such as \textit{fearful} and \textit{abusive}, the use of an LLM with LoRA proves advantageous. Nonetheless, opportunities for further improvements remain, particularly in the recognition of \textit{abusive} and \textit{dissatisfied}, two crucial emotional states. \\

\begin{table}[b]
\centering
\resizebox{9cm}{!}{%
\begin{tabular}{lccccccc}
\toprule
Model & Inform  & Success & JGA & CBE & Unique tri. & BLEU \\
\midrule
SIMPLE-gpt & 81.98 & 75.72 & 65.06 & \textbf{1.64}* & \textbf{2336.8} & 22.29 \\
EMO-gpt & 82.12 & 76.18 & 64.71 & 1.61 & 2315.8 & 22.4 \\
PREV-gpt & \textbf{83.56}* & \textbf{78.04}* & \textbf{65.21} & 1.59 & 2291.6 & \textbf{22.55}* \\

\midrule
SIMPLE-llama & 78.5 & 70.46 & \textbf{64.2} & 1.95 & 4054.8 & 22.78 \\
EMO-llama & 78.36 & 70.28 & 64.0 & \textbf{1.96} & 4151.2  & \textbf{22.91} \\
PREV-llama & \textbf{83.32}* & \textbf{75.14}*  & 63.08 & \textbf{1.96} & \textbf{4309.0}*  & 22.35 \\
\bottomrule
\end{tabular}
}
\caption{\small Mean MultiWOZ metrics (5 seeds) and lexical diversity metrics. Best values are in bold and are computed for each respective language model. * indicates statistical significance ($p<$0.05, paired t-test) between best and second best values.}
\label{tab:tod_res}
\end{table}

With regards to \textbf{TOD metrics} (Table \ref{tab:tod_res}), we compare our variants to a SimpleTOD baseline (SIMPLE) trained with each respective language model. For both GPT-2 and Llama-2,  the incorporation of ED into the pipeline yields improved or comparable results. Notably, the inclusion of emotions does not adversely affect any metric with statistical significance, while both success and inform rates see improvements for the PREV variant. This indicates that employing this approach helps to more closely replicate reference dialogue acts and responses, which largely determine  these improved metrics.

On the other hand, the EMO variant does not offer as much signal, yielding results similar to the baseline in each case.  Compared to GPT-2, Llama-2 responses exhibit greater diversity, as indicated by the CBE score and the unique trigram count, while simultaneously aligning with the gold responses, as reflected by BLEU scores. However, inform and success rates are slightly lower than those of PREV-gpt. 

Overall, these results validate the viability of unifying ED and TOD modeling. They suggest that incorporating user emotions introduces minimal noise and, in fact, contributes meaningful signal.

\begin{table}[!t]
    \centering
    \resizebox{7cm}{!}{
    \begin{tabular}{cccccc}
    \toprule
         Model &  \#1 &  \#2&   \#3 & Mean Rank & $\kappa$\\
         \midrule
         SIMPLE & 35.56\%  & 40.56\%  & \textbf{23.89}\%  & 1.88   & 0.57\\
         PREV & 40.0\%  & \textbf{51.67}\%  & 8.33\%  & 1.68 & 0.43\\
         
        REFINE & \textbf{70.0}\%  & 22.22\%  & 7.78\%  & \textbf{1.38}  & 0.41\\
 \bottomrule
    \end{tabular}
    }
    \caption{\begin{small}Mean human rankings of responses from three Llama-2-based variants, following an emotional user utterance. The first three columns represent the distribution of rankings per model in \%. Rank 1 is best, so a lower rank implies preferred responses. We compute Fleiss's $\kappa$ to characterize agreement on the ranks attributed to each model.\end{small}}
    \label{tab:human_ranking}
\end{table}

\paragraph{\textbf{Human Evaluation}}
To assess the impact of conditioning responses on explicit emotions, we conduct an in-house human evaluation using delexicalized Llama-2-based responses (Table \ref{tab:human_ranking}). Our comparison involves SIMPLE baseline responses, PREV responses conditioned on user emotions through joint training, and REFINE responses, which are PREV responses more explicitly conditioned via prompting. We adopt a methodology inspired by \cite{nekvinda-dusek-2022-aargh} which involves side-by-side relative ranking, randomly selecting examples from each emotional category (\textit{neutral} excluded). We create a collection of 60 evaluation examples, annotated by three distinct NLP practitioners.

For each example, participants are provided with the Context, the reference emotion label for the last user utterance and three responses to rank from best to worst. We allow the same rating to be applied to multiple responses if annotators perceive them as highly similar. Annotators are explicitly instructed to rate responses based on how well they account for the user's emotion and on how well they assist the user in completing their booking. Our annotation interface is shown in Appendix 1.

Our findings reveal a preference for REFINE and PREV responses, compared to baseline SIMPLE responses.  REFINE especially demonstrates a more consistent understanding of the user's emotional state and drives engagement by making the user feel heard (Table \ref{tab:response_examples}). However, despite REFINE achieving a mean rank closer to 1, we identify instances where its responses are less coherent than those from the PREV variant. This discrepancy arises when the additional segments are redundant or confusing, which is sometimes caused by inaccuracies in emotion prediction. PREV proves more resilient to such errors, although it generally displays less emotion awareness. Annotators also at times observe minimal variation between responses, suggesting that while our enhanced variants stay on track with respect to the task, there is room for improvement in showcasing empathy.

\section{Related work} \label{sec: related_work}
Work relating to customer support dialogues has also focused on the intersection between ED and TOD, as these exchanges often stem from a poor user experience, making ED key. For instance, \cite{devillers-2003} annotate emotions in a small corpus of 100 spoken call-center dialogues.  \cite{zuters2020adaptive} also provide valuable insight, examining emotion intensity and frustration in customer support dialogues. However, the datasets utilized offer limited task-oriented annotations and lack explicit ties to a database, making it challenging to experiment with typical multi-domain TOD system baselines. \cite{wei2021target} demonstrate a similar interest as ours by endeavoring to create emotion-aware responses. In contrast to our work however, they assume the user's emotion is given. Additionally, they focus on a dataset of general social media post/response pairs, very different from the TODs to be found in MultiWOZ. \cite{feng-etal-2023-chatter} enhance emotion recognition models for chitchat and apply them to the complete EmoWOZ dataset. The full dataset contains an extra 1000 human-machine dialogues, not annotated with task information.  Due to our specific approach, we focus solely on experimenting with the MultiWOZ portion of EmoWOZ and do not compare ED results with this study. Other research explores the use of emotionally-grounded user simulators for training dialogue policy models through reinforcement learning \cite{lin2023emous}. In this setting, however, the system is not expected to predict the associated emotion label, in contrast to our endeavor.


\begin{table}[!t]
    \centering
    \resizebox{15cm}{!}{
    \begin{tabular}{p{3.0in}|p{4.0in}}
    \toprule
      \textbf{\textit{Dialogue History}}   &  \textbf{\textit{Responses}} \\
         \midrule
          \textbf{1.} & \textbf{SIMPLE}: The address is [address]  and the phone number is [phone].\\
          (...) \textit{User}: Can you tell me the address to the police station? I am in trouble and I need some help asap. (\textit{fearful}) & \textbf{PREV}: The police station is located at [address].
         
\textbf{REFINE}: Oh no, I'm so sorry to hear that! The police station is located at [address].\\
        
        \midrule
        
         \textbf{2.} & \textbf{SIMPLE}:  I have [choice]  options for you. [name] , [name] , and [name] . Do any of these sound good? \\
         (...) \textit{User}: It's our anniversary and my wife loves british food. I want it to be special, so maybe something expensive. (\textit{excited}) & \textbf{PREV}: I have [name]  in the [area]  that serves [food]  food and is in the [pricerange]  price range.\\ 
        
        & \textbf{REFINE}: Of course! I'd be delighted to help you find the perfect place for your anniversary! I have [name]  in the [area]  that serves [food]  food and is in the [pricerange]  price range.\\

         \midrule
         \textbf{3.} & \textbf{SIMPLE}:  The postcode for [name]  is [postcode]  and they do have free parking. \\
         (...) \textit{User}: You didn't answer my question. I need the postcode for The Cambridge Belfry and I need to know if they have free parking or not. (\textit{abusive}) & \textbf{PREV}: I'm sorry about that. The postcode for [name]  is [postcode] . They do have free parking. Is there anything else I can help you with?
        
\textbf{REFINE}: I apologize, I did not see the question. I'm sorry about that. The postcode for [name]  is [postcode] . They do have free parking. Is there anything else I can help you with?\\

        \midrule
        \textbf{4.}  & \textbf{SIMPLE}: Thank you for using our service. \\
        (...) \textit{System}: Is there anything else you need help with today? & \textbf{PREV}: Thank you for using our service. \\
        
        \textit{User}: That shall be all. Thanks! (\textit{satisfied}) & \textbf{REFINE}: Yes you did! You're very welcome! Thank you for using our service.\\
        \bottomrule 
    \end{tabular}
    }
    \caption{\begin{small} Examples of delexicalized system responses following emotional user turns (dialogues pmul3748, pmul4894, pmul4186 and sng1016). Examples 1. and 2. demonstrate how REFINE can successfully add emotion-awareness to responses. Example 3. and 4. respectively show how the added snippets can however be slightly redundant or confusing at times. Example 3. also shows PREV providing adequate emotion awareness compared with SIMPLE.\end{small}}
    \label{tab:response_examples}
\end{table}

\section{Conclusion}
Our findings demonstrate that unifying ED and TOD modeling is overall beneficial for both tasks, producing improved results across ED and task-oriented metrics.  Additionally, our experiments reveal that explicitly predicting emotion labels provides useful grounding for system responses, resulting in a heightened display of empathy especially when refining responses. In a nutshell, our findings underscore the advantages and ease of integrating ED into TOD modeling, compared with treating it as an independent task. We believe our contribution will stimulate TOD system developments which extend beyond the sole objective of task completion, fostering more user-focused advancements benefiting various conversational settings \cite{campillos2020designing, kim2004effects}.

\section{Acknowledgments}
This work was granted access to the HPC resources of IDRIS under the allocation 20XX-AD011014510 made by GENCI.

\section*{Appendix 1: Human Evaluation Interface} \label{app: interface}
\begin{figure}[b]
\centering
\includegraphics[width=15.3cm]{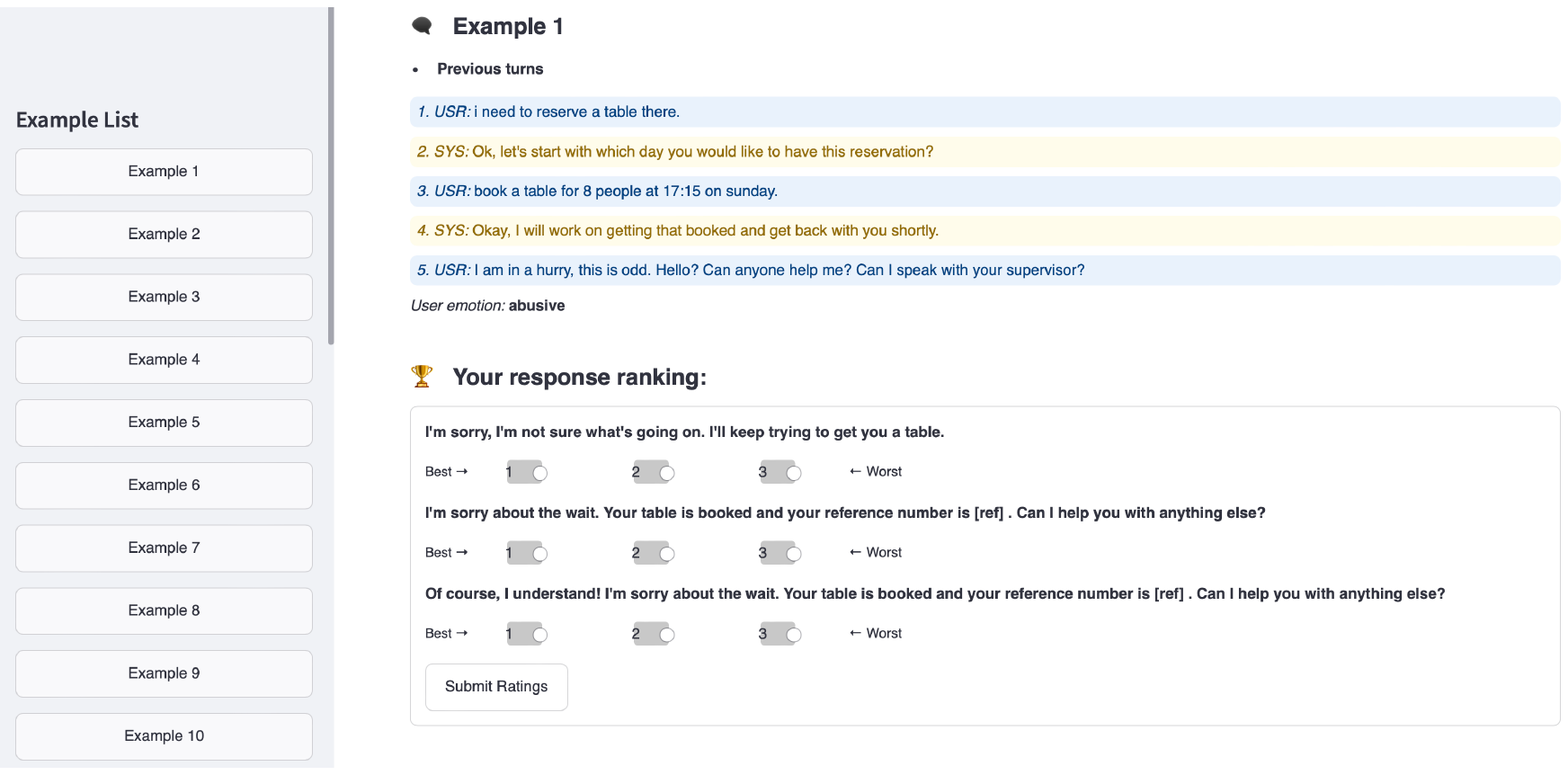}
\caption{Our graphical user interface used for human evaluations.}
\label{fig:interface}
\end{figure}

We adapt the graphical user interface from \cite{nekvinda-dusek-2022-aargh} to fit our particular needs (Figure \ref{fig:interface})

\section*{Appendix 2: Training and Implementation Details} \label{app: training_details}
We train  models with the huggingface\footnote{https://huggingface.co/} and pytorch frameworks. For each model, we perform grid search over learning rates ranging from 1e-5 to 2e-4 with an effective batch size of 32 (we use batches of 16 and 2 gradient accumulation steps). We use the validation set to pick the most effective learning rates. When training GPT-2, we use a learning rate of 8e-5 and train the model for up to 10 epochs with early stopping.  When performing LoRA fine-tuning, we choose the following LoRA configuration: a rank of size 32, given the complexity of our task, a scaling $\alpha$ of 32, and target modules which include the query and value projection matrices, found in the self-attention module of each transformer block. We use a learning rate of 5e-5 for SIMPLE and EMO and 4e-5 for PREV, training for up to 3 epochs with early stopping. Training and inference is done on a single 80Gb A100 GPU using 5 random seeds for each experiment. We generate outputs using greedy decoding and prevent the models from repeating n-grams of size 10, to avoid repetitive loops that may arise when using greedy search. For prompting, we use sampling with a temperature of 0.9.

\section*{Appendix 3: REFINE Prompt Template} \label{app: template}
We use a few-shot chain-of-thought template to prompt Llama-2. Exemplars follow the template below, which include a \textit{Thought} component highlighting the behavior to exhibit in the generated snippet.  We concatenate one exemplar per emotion, excluding \textit{neutral}. 
\begin{small}
\begin{verbatim}
Instruction: 
In the provided dialogues, you'll find conversation contexts 
involving a user and an information system that assists with 
various tasks. The user's emotion is indicated. If the original 
response does not address the user's emotion, your task is to write 
a brief snippet that can be added BEFORE the original response. 
The snippet should convey remorse, reassurance, appeasement, or 
enthusiasm, depending on the user's emotion. Refrain from repeating 
previous snippets or the original response.
------------
Context:
<|user|> While in Cambridge I need a restaurant that is moderately 
    priced and serves Spanish food.
<|system|> La Tasca would be a good choice for you. Would you like 
    the address or phone number?
<|user|> Yes I would like the address and phone number, thanks for 
    your help!
<|system|> The address for La Tasca is 14 -16 Bridge Street and the 
    phone number is 01223464630. Would you like for me to book you 
    a table?
<|user|> I just want to confirm. This place has free wifi as well as 
    being 3 star, correct?
<|system|> Were you also trying to book a hotel today? There are 5 
    hotels in cambridge that are 3 stars and offer free wifi
<|user|> Actually, I'd like to confirm that La Tasca is a 3-star 
    restaurant.
<|system|> Thank you. Have a good day!
<|user|> Wait please you haven't answered my question.

User is feeling: 
dissatisfied, disliking

Original Response: 
<|system|> [name] is the restaurant and is not assigned a star rating

Thought: 
Displaying remorse would be appropriate.
            
Add before the original response: 
Oh I'm sorry!
------------
(...)
\end{verbatim}
\end{small}

\section*{Appendix 4: Emotion Label Examples from EmoWOZ}
 EmoWOZ user utterances and their associated emotion labels can be found in Table \ref{tab:label_examples}. We direct readers to the paper for more details.

\begin{table}[!h]
    \centering
    \resizebox{14cm}{!}{
    \begin{tabular}{p{1.5in}|p{4.0in}}
    \toprule
       
      \multicolumn{1}{c}{\textbf{\textit{Emotion}}}   &   \multicolumn{1}{c}{\textbf{\textit{Example}}}\\
         \midrule

          \textbf{Neutral} & \textit{System}: What sort of food would you like it to be?\\
           & \textit{User}: You choose. Book me a table for 6 people at 12:00 on Thursday.\\

           \midrule

           \textbf{Fearful, sad, disappointed} & \textit{User}: Please help I've been robbed!! \\
        
        \midrule
            
           & (Explicit expression) \\
           & \textit{User}: Actually, could you get me some information about museums in the area? \\
           & \textit{System}: We actually have 23 restaurants in town! Is there a certain area you'll be in? \\
           & \textit{User}: I'm not looking for a restaurant, I want a museum !\\
            \cline{2-2}
        \textbf{Dissatisfied, disliking}& (Implicit expression) \\
            & \textit{User}: I need a taxi from the hotel to the museum after 23:45.\\
            & \textit{System}: Do you want the hotel reservations to begin on Monday? \\
            & \textit{User}: We're talking about a taxi now.\\
            & \textit{System}: You would love broughton house gallery.\\ 
            & \textit{User}: Taxi.\\

        \midrule

         \textbf{Apologetic} & \textit{System}: There are several Italian Restaurants. What area are you interested in?\\
           & \textit{User}: I am so sorry, I would rather have gastropub food, moderately priced please.\\

         \midrule
          & \textit{System}: ...There are 10 attractions in the east - do you care what sort?\\
         \textbf{Abusive}  & \textit{User}: Just do what I say! I don't care what you choose! Give me the necessary information too.\\

          \midrule
         \textbf{Excited, happy, anticipating} & \textit{User}: I'm traveling to Cambridge and looking forward to their restaurants ! I need a place to stay while I'm there.\\

          \midrule
         \textbf{Satisfied, liking, appreciative} & \textit{System}: Is there anything else I can assist you with?\\
           & \textit{User}: No, thanks you've been great!\\

         \bottomrule
        
    \end{tabular}
    }
    \caption{\begin{small} Examples for each emotion label from EmoWOZ \cite{feng-etal-2022-emowoz}.\end{small}}
    \label{tab:label_examples}
\end{table}

\newpage

\printbibliography

\end{document}